\documentclass[10pt,twocolumn,letterpaper]{article}

\usepackage{cvpr}              

\usepackage[super]{nth}
\usepackage[accsupp]{axessibility}
\usepackage{amsmath}
\makeatletter
\robustify\@latex@warning@no@line
\makeatother
\usepackage{authblk}

\definecolor{cvprblue}{rgb}{0.21,0.49,0.74}
\usepackage[pagebackref,breaklinks,colorlinks,citecolor=cvprblue]{hyperref}

\def\paperID{00074} 
\def\confName{CVPR}
\def\confYear{2025}

\title{NTIRE 2025 Challenge on Video Quality Enhancement for Video Conferencing: Datasets, Methods and Results}

\author{
	{Varun Jain}$^{1}$, {Zongwei Wu}$^{2}$, {Quan Zou}$^{1}$, {Louis Florentin}$^{1}$, {Henrik Turbell}$^{1}$, {Sandeep Siddhartha}$^{1}$,
    
    {Radu Timofte}$^{2}$,
    {Qifan Gao}$^{*}$, {Linyan Jiang}$^{*}$, {Qing Luo}$^{*}$, {Jack Song}$^{*}$, {Yaqing Li}$^{*}$, 
    
    {Summer Luo}$^{*}$, 
    {Mae Chen}$^{*}$, {Stefan Liu}$^{*}$, {Danie Song}$^{*}$, {Huimin Zeng}$^{*}$, {Qi Chen}$^{*}$,
    
    {Ajeet Verma}$^{*}$, {Shweta Tripathi}$^{*}$, {Vinit Jakhetiya}$^{*}$, {Badri N Subhdhi}$^{*}$, {Sunil Jaiswal}$^{*}$
    \\    
    $^{1}${Microsoft} and $^{2}${University of W\"urzburg} \\

    {\tt\small jain.varun@microsoft.com, zongwei.wu@uni-wuerzburg.de, \{quan.zou, lflorentin, heturbel, ssiddhartha\}@microsoft.com, radu.timofte@uni-wuerzburg.de}
}

\begin{document}
    \maketitle
    {\let\thefootnote\relax\footnotetext {
    ${^*}$ These members were participants who co-authored this report detailing their methodologies, not the challenge organizers. Please refer to Appendix~\ref{sec:appendix_teams} for their correspondence details.
} }

\begin{abstract}
\label{sec:abstract}

This paper presents a comprehensive review of the \nth{1} Challenge on Video Quality Enhancement for Video Conferencing held at the NTIRE workshop at CVPR 2025, and highlights the problem statement, datasets, proposed solutions, and results. The aim of this challenge was to design a Video Quality Enhancement (VQE) model to enhance video quality in video conferencing scenarios by (a) improving lighting, (b) enhancing colors, (c) reducing noise, and (d) enhancing sharpness –- giving a professional studio-like effect.
Participants were given a differentiable Video Quality Assessment (VQA) model, training, and test videos. A total of 91 participants registered for the challenge. We received 10 valid submissions that were evaluated in a crowd-sourced framework. Additional materials can be found on the project website~\footnote{\url{https://www.microsoft.com/en-us/research/academic-program/ntire-2025-vqe/}}\textsuperscript{,}\footnote{\url{https://github.com/varunj/cvpr-vqe/}}.

\end{abstract}

    \section{Introduction}
\label{sec:intro}

Light is a crucial component of visual expression and is the key to controlling texture, appearance, and composition. Professional photographers often have sophisticated studio lights and reflectors to illuminate their subjects such that the true visual cues are expressed and captured. Similarly, tech-savvy users with modern desk setups employ a sophisticated combination of key and fill lights to give themselves control over their illumination and shadow characteristics.
However, many users are constrained by their physical environment, which may lead to poor positioning of ambient lighting or lack thereof. It is also commonplace to encounter flares, scattering, and specular reflections that may come from windows or mirror-like surfaces. Problems can be compounded by poor-quality cameras that may introduce sensor noise. This leads to poor visual experience during video calls and can have a negative impact on downstream tasks such as denoising, super-resolution, segmentation, and face detection.

The current light correction solution in Microsoft Teams, called AutoAdjust, finds a global mapping of input to output colors which is updated sporadically. Since this mapping is global, it gives more importance to foreground colors, which may lead to improper exposure of or color shifts in the background.
On the other hand, popular single-image portrait relighting methods~\cite{zhou2019deep} estimate local correction in only the foreground and preserve the background by an implicit in-network matte layer. A possible side effect of local correction can be the reduction of local contrast, which often serves as a proxy to convey depth in 2D images, making people appear dull in some cases.

We conducted P.910~\cite{naderi2024crowdsourcing} studies totaling 350,000 pairwise comparisons that measured people’s preference for AutoAdjust and portrait relighting over no effect and images manually edited by experts
in Adobe Lightroom. We used the Bradley–Terry model~\cite{bradley1952rank} to estimate the scores for each method and observed that people preferred AutoAdjust more than any other method.

\begin{figure*}[ht]
    \centering
    \includegraphics[width=\linewidth]{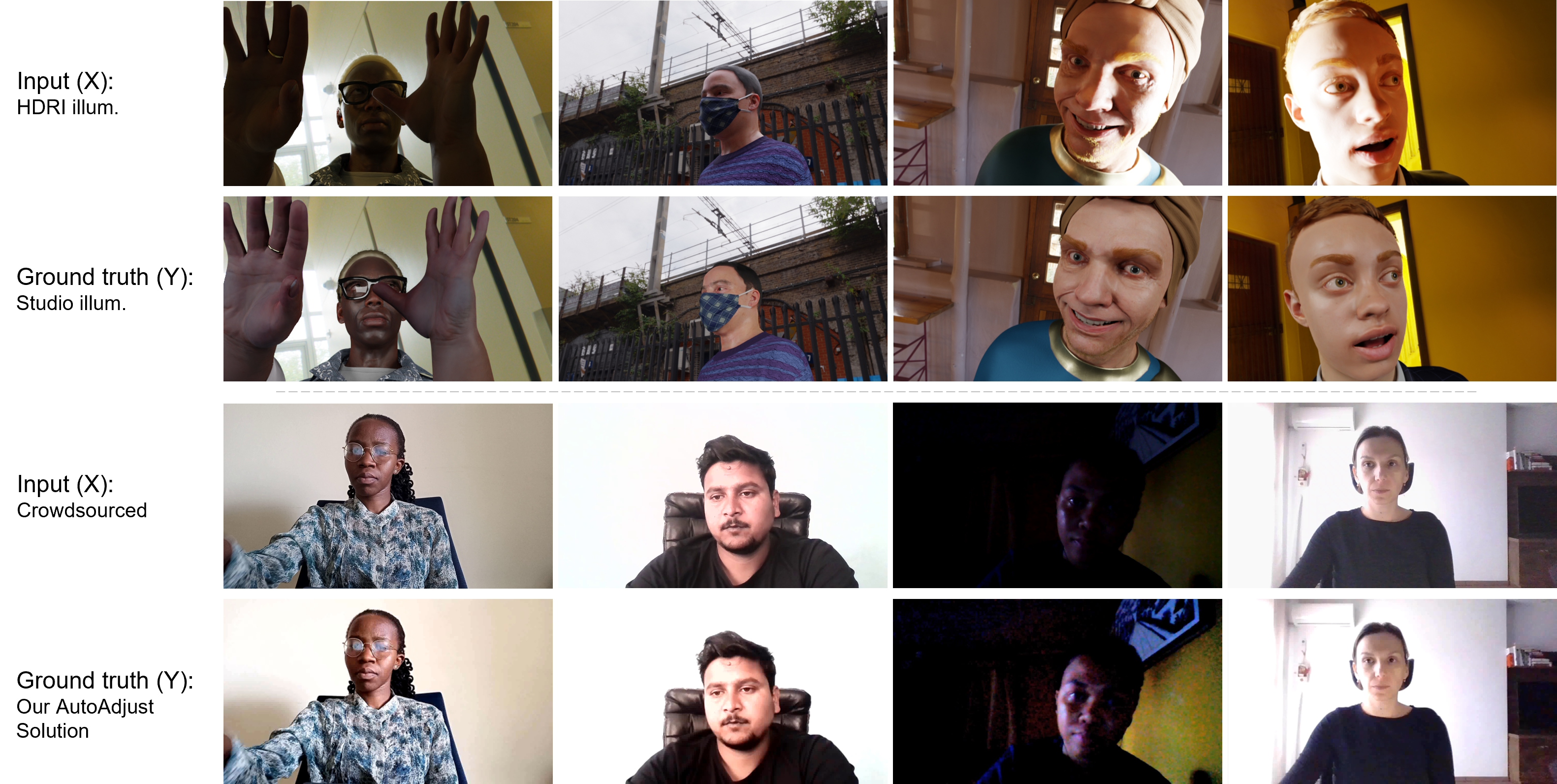}
    \caption{Ground truth from (top) our synthetics framework, (bottom) the AutoAdjust solution. The top row shows the input with suboptimal foreground illumination which is fixed by adding a studio light setup in front of the subject which is simulated in synthetics and predicted via global changes in the real data.}
    \label{fig:data}
\end{figure*}

To take the next step towards achieving studio-grade video quality, one would need to (a) understand what people prefer and construct a differentiable Video Quality Assessment (VQA) metric, and (b) be able to train a Video Quality Enhancement (VQE) model that optimizes this metric. To solve the first problem, we have trained a VQA model that, given a pair of videos $x_1$ and $x_2$, gives the probability that $x_1$ is better than $x_2$ as described in Sec.~\ref{sec:challenge_vqa_model}. Given a standard test set, this information can be used to construct a ranking order of a given set of methods.

We invited researchers to participate in a challenge aimed at developing Neural Processing Unit (NPU) friendly VQE models that leverage our trained VQA model to improve video quality.

This challenge was one of the NTIRE 2025~\footnote{\url{https://www.cvlai.net/ntire/2025/}} Workshop associated challenges on: ambient lighting normalization~\cite{ntire2025ambient}, reflection removal in the wild~\cite{ntire2025reflection}, shadow removal~\cite{ntire2025shadow}, event-based image deblurring~\cite{ntire2025event}, image denoising~\cite{ntire2025denoising}, XGC quality assessment~\cite{ntire2025xgc}, UGC video enhancement~\cite{ntire2025ugc}, night photography rendering~\cite{ntire2025night}, image super-resolution (x4)~\cite{ntire2025srx4}, real-world face restoration~\cite{ntire2025face}, efficient super-resolution~\cite{ntire2025esr}, HR depth estimation~\cite{ntire2025hrdepth}, efficient burst HDR and restoration~\cite{ntire2025ebhdr}, cross-domain few-shot object detection~\cite{ntire2025cross}, short-form UGC video quality assessment and enhancement~\cite{ntire2025shortugc,ntire2025shortugc_data}, text to image generation model quality assessment~\cite{ntire2025text}, day and night raindrop removal for dual-focused images~\cite{ntire2025day}, video quality enhancement for video conferencing, low light image enhancement~\cite{ntire2025lowlight}, light field super-resolution~\cite{ntire2025lightfield}, restore any image model (RAIM) in the wild~\cite{ntire2025raim}, raw restoration and super-resolution~\cite{ntire2025raw} and raw reconstruction from RGB on smartphones~\cite{ntire2025rawrgb}.

    \section{Challenge}
\label{sec:challenge}

\subsection{Problem Statement}

The task was to enhance video quality in video conferencing scenarios. We only looked at the following properties of a video to judge its studio-grade quality:

\begin{enumerate}
    \item Foreground illumination – the person (all body parts and clothing) should be optimally lit.
    \item Natural colors – correction may make local or global color changes to make videos pleasing.
    \item Temporal noise – correct for image and video encoding artefacts and sensor noise.
    \item Sharpness - to ensure that correction algorithms do not introduce softness, the final image should at least be as sharp as the input.
\end{enumerate}

We understand that there may be many other aspects to a good video. For simplicity, we discounted all except the ones mentioned above. Specifically, we did not measure the following:

\begin{enumerate}
    \item Egocentric motion – unstable camera may introduce sweeping motion or small vibrations that we did not aim to correct.
    \item Makeup and beautification – it is commonplace for users to apply beautification filters that alter their skin tone and facial features such as those found on Instagram and Snapchat. We did not aim for that aesthetic.
    \item Removal of reflection on glasses and lens flare - although it is a common occurrence in video teleconference scenarios, we did not aim to remove reflections that may come from screens and other light sources onto users’ glasses due to the risk associated with altering eye appearance and gaze direction.
    \item Avatars – A solution that synthesizes a photorealistic avatar of the subject and drives it based on the input video would score the highest in terms of noise, illumination, and color. If it indeed minimizes the total cost function that takes into account all these factors, it would be acceptable.
\end{enumerate}

\subsection{Baseline Solution}

Since the AutoAdjust model was ranked higher than expert-edited images and portrait relighting methods, we provided participants a baseline solution so that they could reproduce the AutoAdjust feature as currently shipped in Microsoft Teams. It was provided as a Python script that calls the AutoAdjust executable, and includes code for post-processing.

\subsection{Compute Constraints}

The goal was to have a computationally efficient solution that can be offloaded to NPU for CoreML inference. We established a qualifying criterion of CoreML uint8 or fp16 models with at most $20.0$x$10^9$ MACs per frame for an input resolution of $1280 \times 720$. We estimate such a model to have a per frame processing time of $9$ ms on an M1 Ultra powered Mac Studio and $5$ ms on an M4 Pro powered Mac Mini for the given input resolution. Submissions that did not meet this criterion were not considered for the P.910 evaluation.

\begin{table*}[tp]
    \centering
    \resizebox{\textwidth}{!}{
    \begin{tabular}{c|c|c|c|c|c|c|c|c|c|c}
        \hline
        Teamname & Input Resolution & Inference Resolution & Training Time & Epochs & Ensemble & LUT & Attention & \#MACs/frame & Latency/frame & GPU/NPU  \\
        \hline
        TMobileRestore & (720, 1280, 3) & (720, 1280, 3) & 1 day & 100 & Yes &  Yes & No & $16.8 \times 10^9$ & 200ms & V100\\
        \hline
        Summer & (720, 1280, 3) & (720, 1280, 3) & 2 days & 120 & Yes &  Yes & No & $15.4 \times 10^9$ & 180ms & V100\\
        \hline
        XTeam & (720, 1280, 3) & (720, 1280, 3) & 6 hrs & 25 & Yes &  Yes & No & $16.8 \times 10^9$ & 200ms & V100\\
        \hline
        Velta & (720, 1280, 3) & (720, 1280, 3) & 12 hrs & 30 & Yes &  Yes & No & $15.4 \times 10^9$ & 180ms & V100\\
        \hline
        DeepView & (720, 1280, 3) & (720, 1280, 3) & 7 days & 80 & Yes &  No & Yes & $106.4 \times 10^9$ & 238ms & V100\\
        \hline
        Auv & (720, 1280, 3) & (720, 1280, 3) & 1 day & 5 & Yes &  No & Yes & $106.4 \times 10^9$ & 238ms & V100\\
        \hline
        Meeting & (720, 1280, 3) & (720, 1280, 3) & 3 hrs & 10 & Yes &  Yes & No & $16.8 \times 10^9$ & 200ms & V100\\
        \hline
        Maqic & (720, 1280, 3) & (720, 1280, 3) & 7 days & 157 &  No &  Yes & No & $13.0 \times 10^3$ & 28 ms & V100 \\
        \hline
        LUT & (720, 1280, 3) & (720, 1280, 3) & 4 days & 80 &  No &  Yes & No &  $13.0 \times 10^3$ & 27 ms & V100 \\
        \hline
        Wizard & (720, 1280, 3) & (720, 1280, 3) & 15 days & 50 & Yes &  No & Yes & $114.2 \times 10^9$ & 170ms & V100\\
        \hline
    \end{tabular}
    }
    \caption{Final results of the NTIRE 2025 Challenge on Video Quality Enhancement for Video Conferencing held at CVPR 2025.}
    \label{table:results}
\end{table*}

\subsection{Dataset}

\begin{figure}[ht]
    \centering
    \includegraphics[width=0.8\columnwidth]{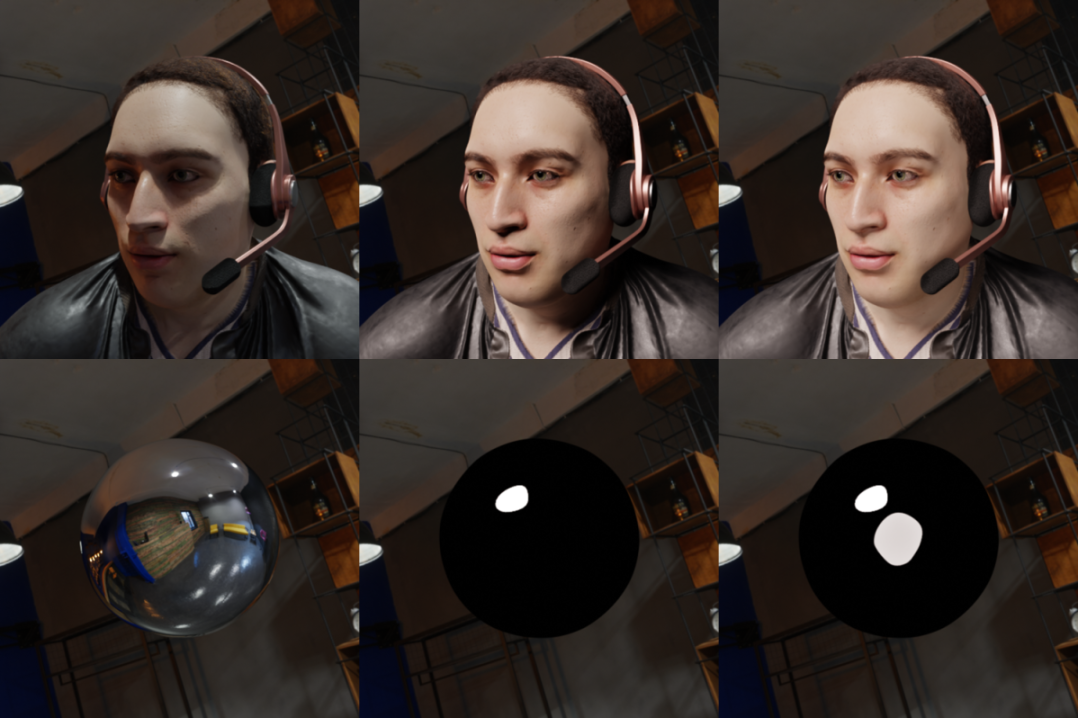}
    \caption{Comparison of lighting setup in the Synthetic Portrait Relighting dataset. (left) Lighting from the HDRI, (center) key light with HDRI lighting turned off, and (right) key and fill lights with HDRI lighting turned off. Note that the HDRI is only used as a background when using the studio lighting and does not contribute to the illumination of the subject.}
    \label{fig:studiolights}
\end{figure}

\begin{figure}[ht]
    \centering
    \includegraphics[width=0.8\columnwidth]{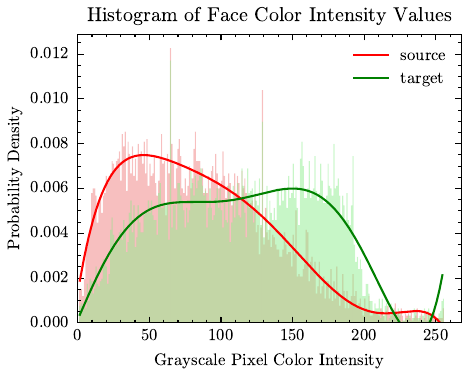}
    \caption{Color intensity in source and target images of our Synthetic Portrait Relighting dataset. The source images are dark with intensity centered around $50$. The target fixes this by boosting illumination -- making it more uniform and span a larger range.}
    \label{fig:facevalues}
\end{figure}

\subsubsection{Unpaired Real Data}
We host a web service that reaches users all over the world and prompts them to sit in front of a laptop or a PC. We then record minute-long videos while users perform hand gestures and body movements. We sampled $13,000$ videos from this dataset for training, validation, and testing of VQE methods. The videos are $10$~s long, encoded at $19$ FPS on average, and amount to a total of $3,900,000$ frames. We kept $3,000$ ($23\%$) videos for testing and ranking submissions and make $10,000$ ($77\%$) available to the teams. They could choose to split it between the training and validation sets as they desire. The teams were also free to use other publicly available datasets, while being mindful about data drift.

Of the $13,000$ videos, we selected $300$ high quality videos where P.910 raters voted strongly in favor of the AutoAdjust result, as shown in the bottom half of \figurename{~\ref{fig:data}}. We assumed these to be the ground truth. P.910 done on these videos shows a Mean Opinion Score (MOS)~\cite{naderi2024crowdsourcing} of $3.58$ in favor of the target.

\subsubsection{Paired Synthetic Portrait Relighting Data}
In addition to these data, we also provided paired data for fully supervised learning as shown in the top half of \figurename{~\ref{fig:data}}. 
Note that it is possible to learn a correction which is different from these ground truth labels and achieve a higher MOS. Hence, these labels had to be treated as suggestive improvements, and not as global optima.

We use a physically-based path tracer and photorealistic assets in Blender to render $1,500$ videos for training and $500$ videos for testing. Each video is $5$~s long encoded at $30$~FPS. The source image has lighting only from the High Dynamic Range Image (HDRI) environment. For the target, we added $2$ diffuse light sources to simulate a studio lighting setup. Refer \figurename{~\ref{fig:studiolights}} to visualize the effect of these light sources and \figurename{~\ref{fig:facevalues}} for statistics on the color intensity values on the face. These are the same images that were used to finetune the portrait relighting method.

To ensure that these data generalize well in the wild, we refer to the image-level degradations used in Real-ESRGAN~\cite{wang2021real} and applied them to the source image. To simulate out-of-focus blur, we applied generalized Gaussian blur kernels~\cite{liu2020estimating} that have ramp edges and flat top areas -- better modeling the combined effects of lens defocusing and light diffraction. For color noise, we used channel-independent additive Gaussian noise, and gray noise was added by applying the same Gaussian noise to all $3$ channels. Finally, sensor noise was modeled by sampling from a Poisson distribution. Lastly, we applied random resizing and JPEG compression.

P.910 done on these videos shows a MOS of $4.06$ in favor of the target indicating that these make for a better target compared to the baseline AutoAdjust solution. Some examples of these pairs are shown in \figurename{~\ref{fig:data}} and more details about the rendering framework can be found in~\cite{hewitt2024look}.

\subsection{VQA Model}
\label{sec:challenge_vqa_model}

\begin{equation}
    p_{x_1, x_2}, A_{x_1}, A_{x_2} = VQA_\theta(x_1, x_2)
    \label{eq:vqa}
\end{equation}

We provided teams with a pre-trained Siamese~\cite{koch2015siamese} Video Quality Assessment model $VQA_{\theta}$ that was trained on $22,553$ videos and $11$ enhancement models. Ground-truth was collected by prompting human raters with $315,636$ side-by-side video comparisons. For high-level semantic understanding, we used our own models that were pre-trained on a collection of real and synthetic images for the tasks of person segmentation, face quality and image aesthetics. For low-level features such as noise, flicker and video coding artifacts we used the DOVER~\cite{wu2023exploring} model. We took the penultimate feature maps of both models, performed average pooling across temporal and spatial dimensions and concatenated them. We then used a set of projections to predict the final logits.

Given a pair of images or videos $x_1$ and $x_2$, the model predicts the probability of $x_1$ being preferred over $x_2$ in a P.910 study. It also provides $11$ auxiliary scores for each input $A = [a_{1}, a_{2}, ... a_{11}]$ that correspond to factors such as image aesthetic, color harmonization, color liveliness, key-lighting, noise, image composition, face capture quality etc. These are supervised with metrics obtained from publicly available Apple Vision APIs.

\subsection{Metrics and Evaluating Submissions}

\begin{figure*}[htp]
    \centering
    \begin{subfigure}{\textwidth}
        \centering
        \includegraphics[width=\linewidth]{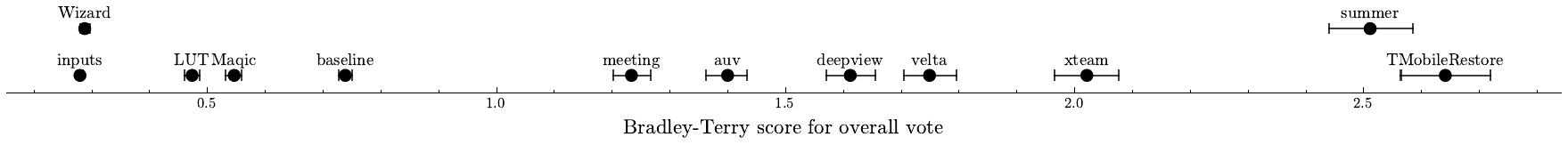}
        \label{fig:results_vote}
    \end{subfigure}
    \hfill
    \begin{subfigure}{\textwidth}
        \centering
        \includegraphics[width=\linewidth]{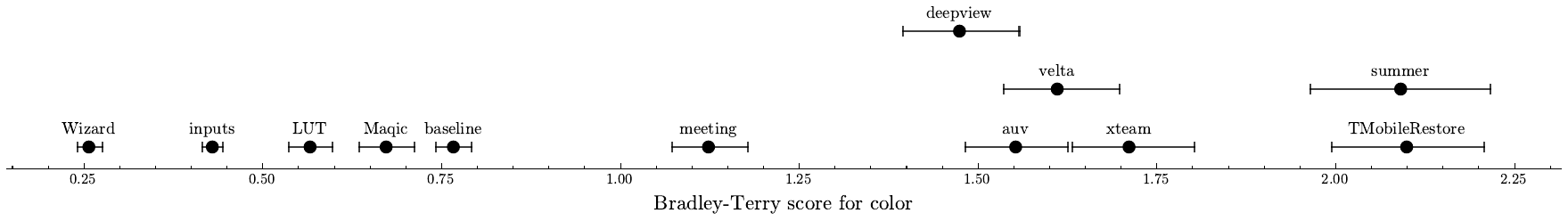}
        \label{fig:results_color}
    \end{subfigure}
    \begin{subfigure}{\textwidth}
        \centering
        \includegraphics[width=\linewidth]{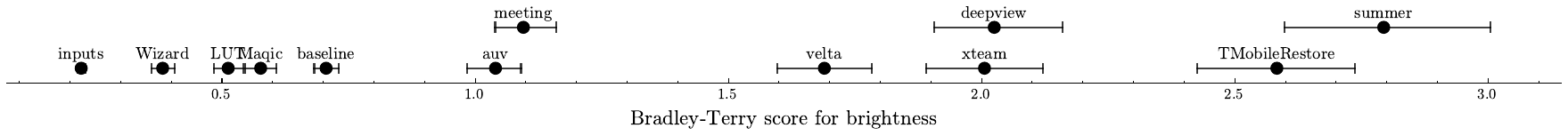}
        \label{fig:results_brightness}
    \end{subfigure}
    \hfill
    \begin{subfigure}{\textwidth}
        \centering
        \includegraphics[width=\linewidth]{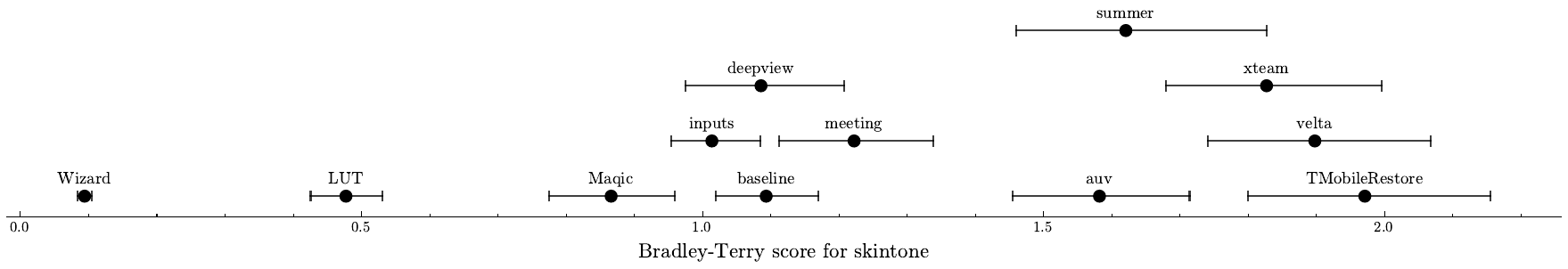}
        \label{fig:results_skintone}
    \end{subfigure}
    \caption{Interval plots illustrating the mean P.910 Bradley-Terry scores and their corresponding $95$\% confidence intervals for the $10$ submissions, input videos, and the provided baseline. (Top) Overall preference, and (bottom) factors influencing preference.}
    \label{fig:results}
\end{figure*}

The final goal was to rank the submissions according to the P.910 scores. We asked the teams to submit their predictions on the $3,000$ real-video test set. We then compared the submissions to the given input, the baseline, and against each other. As shown in \figurename{~\ref{fig:results}}, comparison using the Bradley–Terry model gives us the score for each submission that maximizes the likelihood of the observed P.910 voting. Our P.910 framework has a throughput of $210,000$ votes per week.
In case two methods had statistically insignificant difference in subjective scores, we used the objective score shown in Equation~\eqref{eq:metric} to break ties.

\begin{equation}
    S^{real}_{obj}(\hat{Y},X,\theta) = \frac{1}{12n} \sum_{i=1}^{n} \{{p_{\hat{y_i}, x_i}, A_{\hat{y_i}}}\} | _{VQA_\theta(\hat{y_i}, x_i)}
    \label{eq:metric_real}
\end{equation}
\begin{equation}
    S^{synth}_{obj}(\hat{Y},Y) = \frac{1}{\frac{1}{n} \sum_{i=1}^{n} \sqrt{E[(Y - \hat{Y})^2]}}
    \label{eq:metric_synth}
\end{equation}
\begin{equation}
    S_{obj}(\hat{Y},Y,X,\theta) = S^{real}_{obj}(\hat{Y},X,\theta) \times S^{synth}_{obj}(\hat{Y},Y)
    \label{eq:metric}
\end{equation}

Due to the infeasibility of getting P.910 scores in real-time, teams could use the objective score $S_{obj}$ for continuous and independent evaluation. For the $3,000$ unsupervised videos, teams were required to submit the per-video VQA score $p_{\hat{y_i}, x_i}$ along with the $11$ auxiliary scores $A_{\hat{y_i}}$ predicted by the VQA model as shown in Equation~\eqref{eq:vqa}. For the synthetic test set, the teams reported the Root Mean Squared Error (RMSE) per video. These scores were also published on the leaderboard so that participants could track their progress relative to other teams. However, we did not rank the teams based on these objective metrics since it was possible to learn a correction that is different from and subjectively better than the ground truth provided.

\section{Results}
\label{sec:results}

We received $5$ complete submissions for both the mid-point and final evaluations. For each team’s submission, we utilized our crowd-sourced framework to evaluate their $3,000$-video test set. This involved presenting human raters with $270,000$ side-by-side video comparisons. The raters were asked to provide their preference on a scale of $1$ to $5$, where $1$ and $5$ represent strong preference for the left and right video respectively, and $2$ and $4$ represent weak preference. A rating of $3$ indicates no preference. Furthermore, raters were prompted to specify if their decision was primarily influenced by (a) image colors, (b) image brightness, or (c) skin tone. The Bradley–Terry scores for each team that maximize the likelihood of the observed P.910 voting are shown in \figurename{~\ref{fig:results}}.

    \section{Challenge Methods}
\label{sec:methods}

This section outlines the methodologies and datasets used by the highest-ranking submissions. We observe that Look-Up Table (LUT) based solutions TMobileRestore and DeepView scored the highest. This can be attributed to the efficient, yet temporally stable nature of the correction when compared to methods that predict dense pixel-to-pixel mapping between the input and output image pairs.

\subsection{TMobileRestore}
\begin{figure}[ht]
    \centering
    \includegraphics[width=0.8\columnwidth]{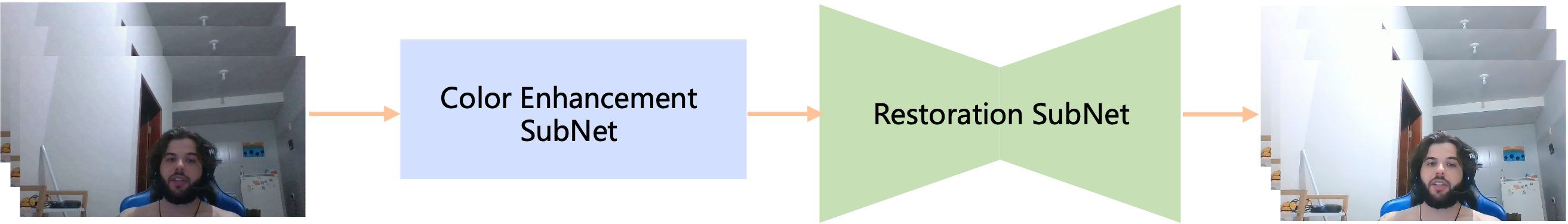}
    \caption{Two stage video conferencing enhancement framework proposed by team TMobileRestore.}
    \label{fig:framework}
\end{figure}

\subsubsection{Description}
They propose a video enhancement algorithm designed to tackle common issues found in video conferencing videos, such as noise, compression artifacts, pathological illumination, and visual inconsistencies. Their algorithm employs a two-stage training process to achieve optimal results: the first stage uses a LUT for brightness and color correction, while the second stage focuses on removing compression noise, sensor noise, and enhancing the overall video quality, as shown in \figurename{~\ref{fig:framework}}.

The first stage of the video enhancement framework is designed to tackle color distortions that are typically found in video conferencing footage, such as inconsistent lighting and color shifts, which considerably lowers visual quality. To effectively rectify these distortions, they use a combination of Clookup table (CLUT) based methods~\cite{zhang2022clut} and convolutional neural network structures. During this phase, the network processes input frames by extracting features at multiple levels, allowing it to concurrently extract image features. They implement a CLUT, where the neural network predicts content-dependent weights from downsampled input to merge basic CLUTs into an image-adaptive one, thereby enhancing the original input image.

The second stage is dedicated to rectifying low-level distortions such as noise and compression artifacts. To address these problems, they utilize a lightweight U-Net architecture with skip connections, specifically engineered for effective and robust restoration. The network extracts features at various scales, enabling it to concurrently address both local artifacts (like blocky compression noise) and global degradations. Skip connections between the encoder and decoder ensure the preservation of fine-grained details throughout the restoration process. The first phase includes $21$ convolutional layers that have the ability to broaden receptive fields and carry out both global and local refinement for image distortions. This allows the network to restore the natural and visually pleasant context throughout the video.

\subsubsection{Datasets}
To train the two-stage network, they used a combination of public datasets, including LDV3~\cite{yang2022aim}, REDS~\cite{liu2022video}, and datasets provided in this competition.
For realistic degradation simulation, they model mixed distortions to create training data that closely resembled real-world scenarios. The training data for the first stage incorporated color distortions, such as random saturation shifts and contrast adjustments. For the second stage, the training data included randomized degradations such as Poisson-Gaussian noise, motion blur, and H.265/H.264 compression. The degradation parameters were dynamically sampled per batch to improve robustness. For both stages, they applied spatial augmentations such as rotation, flipping, and chromatic aberration, as well as temporal jitter techniques such as frame dropping and shuffling to prevent overfitting.

\subsubsection{Experiments \& Results}
In stage 1, the CLUT is trained with a hybrid loss function combining L1 loss and cosine color shift loss, over $200,000$ iterations with a batch size of $32$ and a patch size of $512 \times 512$. The initial learning rate was set to $0.0002$ and halved every $10,000$ iterations, using the Adam optimizer with $\beta_1=0.9$ and $\beta_1=0.99$.  

For the second stage, the sub-network was optimized using a combination of L2 loss, perceptual loss, LPIPS and GAN loss to enhance textures without over-smoothing, over $300,000$ iterations with a batch size of $16$ and a patch size of $512 \times 512$. After pretraining both stages independently, they jointly fine-tuned the network for an additional $20,000$ iterations with a reduced learning rate of $0.00001$.
Details are listed in Table~\ref{table:results}. XTeam and Meeting are similar to this method with early training termination at $50,000$ and $10,000$ iterations respectively.

\subsection{Summer}
This method also consists of the color enhancement sub-network and the video restoration sub-network. The color enhancement network uses a set of five pre-trained 3D LUTs to dynamically adjust the color and tone of video frames in real-time~\cite{zeng2020learning}. These LUTs are trained using the provided supervised VQE dataset, which ensures that each LUT represents a distinct style of color and tone transformation tailored for video content. To predict the optimal combination of these LUTs for each frame, the method employs a convolutional neural network (CNN) with seven convolutional blocks. This CNN extracts global features from the downsampled video frames, capturing essential characteristics that influence the color-enhancement process. By analyzing these features, the network predicts the weights for blending the five 3D LUTs, resulting in a final LUT that is adapted to the specific content of each video frame.

The video restoration sub-network utilizes seven residual blocks that progressively refine video frames, reducing noise, correcting blurriness, and restoring details. This deep learning approach effectively learns to map from degraded images to high-quality images, ensuring clear and detailed video frames.
Velta is similar to this method, with training ending early in $30,000$ iterations.

\subsection{DeepView}
They propose a video conference enhancement network that addresses both degradation distortion repair and color enhancement to ensure high-quality video communication. For distortion repair, it is the same as TMobileRestore. 

For color enhancement, they adopt the HVI-CIDNet approach~\cite{yan2025hvi}, which includes the HVI color space and the CIDNet architecture. The HVI color space minimizes noise and compresses low-light regions, while CIDNet's dual-branch network handles chromatic denoising and brightness enhancement. By processing images in the HVI color space and applying cross-attention mechanisms, this approach restores natural colors and details, providing vibrant and accurate color representation in video conferencing.
Auv is similar to this method, with training ending early in $5,000$ iterations.

\subsection{Maqic}
\begin{figure}[ht]
    \centering
    \includegraphics[width=0.9\columnwidth]{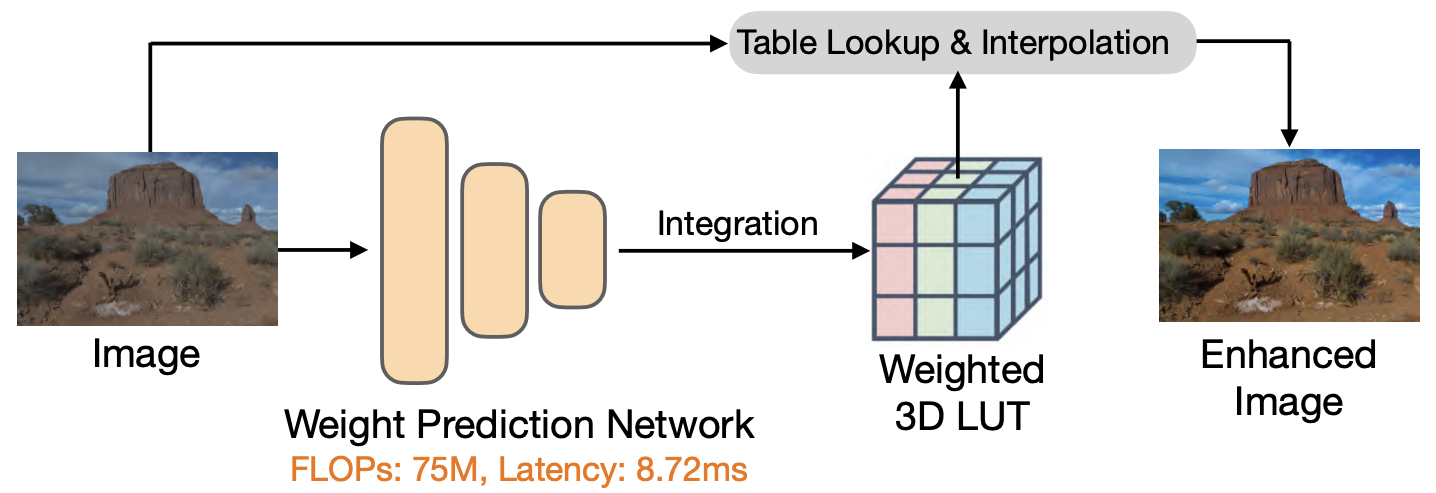}
    \caption{Typical 3DLUT-based retouching pipeline.}
    \label{fig:3dlut}
\end{figure}

\subsubsection{Description}
Online video streams suffer from the physical environment,  including poor positioning of ambient lighting or lack thereof, leading to poor visual experience during video calls and may perturb the downstream tasks. Considering that human-region should be the focus of the video calls, they interpreted this challenge as the task of video portrait retouching, which aims to improve the aesthetic quality of input portrait photos and especially requires human-region priority~\cite{zeng2023region}.  While deep learning-based methods~\cite{ignatov2017dslr,chen2018deep,deng2018aesthetic,zhang2019multiple} largely elevate the retouching efficiency and provide promising retouched results, most of them concentrate on the image tasks, which leads to the efficiency bottleneck when translating to the video task. Therefore, they consider an efficient solution, \ie, a look-up table (LUT) based retouching, which performs fast inverse tone mapping according to the trained look-up table for each pixel value.

\begin{figure}[ht]
    \centering
    \includegraphics[width=0.9\columnwidth]{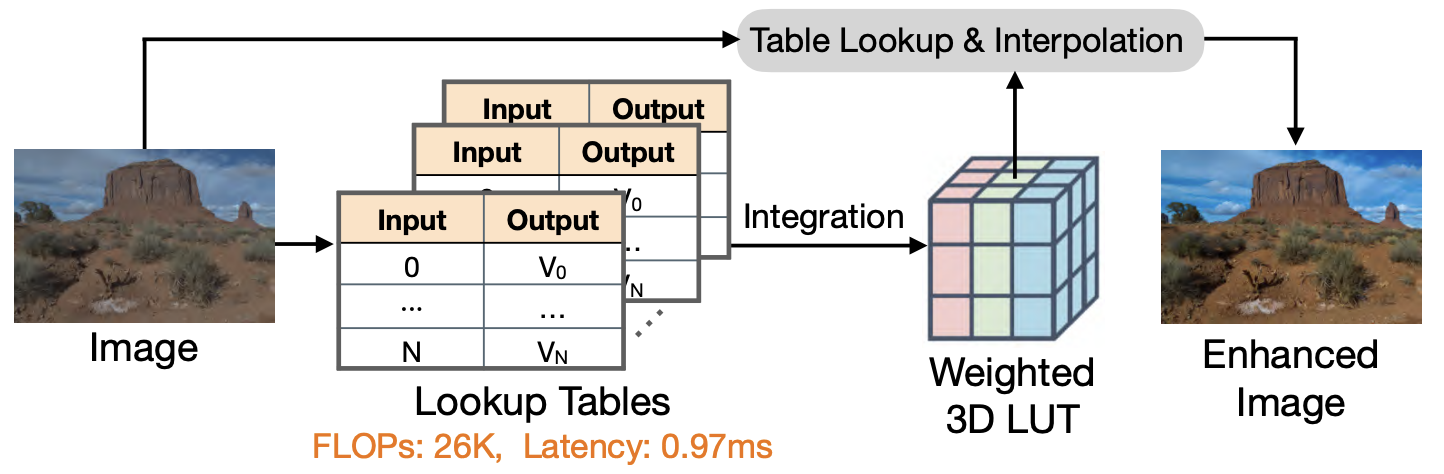}
    \caption{The adopted ICELUT used by Maqic constructs weighted 3D LUT with multiple lookup table candidates to perform real-time video retouching.}
    \label{fig:weighted_3dLUT}
\end{figure}

For the video portrait retouching task, to improve the temporal consistency, existing video-based enhancement methods~\cite{lin2022unsupervised,chan2022basicvsr++} typically include an additional optical flow estimator (\eg, SpyNet~\cite{ranjan2017optical}) to propagate information from adjacent frames. However, this is not suitable for a highly efficient  LUT-based solution, where the online SpyNet inferencing inevitably slows down the whole retouching pipeline. Therefore, they choose the image-based ICELUT~\cite{yang2024taming}  (as shown in \figurename{~\ref{fig:weighted_3dLUT}}) as the retouching backbone. Their contributions can be summarized as choosing an efficient backbone for video portrait retouching and the stage-wise training strategy to achieve perceptual satisfying retouched results.

The typical LUT solution (as shown in \figurename{~\ref{fig:3dlut}}) provides only the LUT-based pixel value transfer to accelerate the retouching process. For the portrait scenario, the retouching should be two-fold: (a) retouching for both the background and foreground, and (b) the focus on highlighting the human region instead of the background. Therefore adopting a solution with a region-wise adaptation (\eg, ICELUT~\cite{yang2024taming}) is necessary to filter out the background and focus on the portrait region. As shown in \figurename{~\ref{fig:weighted_3dLUT}}, given the low-quality input frame, the adopted method adaptively performs the fusion of multiple LUTs and composes the 3D LUT, which then performs tone mapping to obtain the visually satisfying result for each frame.

\subsubsection{Datasets}
They notice that the given dataset contains limited image resolution, with degradations such as noise, motion blur, and flicker. Training with these data complexes the target goal, requiring the model to simultaneously perform retouching and video restoration. Therefore, they adopt a stage-wise training strategy to separate the aforementioned goals.

In the first stage, they train with MIT-Adobe FiveK~\cite{fivek}, which is a high-quality tone mapping dataset without image degradations. This enables the LUTs to retouch input videos, which adaptively changes the tone of frames and adjusts the light condition. Then based on the pre-trained LUTs, they conduct fine-tuning on the supervised subset given in the challenge to equip the LUTs with restoration ability.

\subsubsection{Experiments \& Results}
The model is implemented with the PyTorch framework, they conduct all the experiments on a single NVIDIA Tesla V100 GPU. They include additional details in Table~\ref{table:results}.
LUT is similar to this method with early training termination.

\subsection{Wizard}
\subsubsection{Description}
Their approach~\cite{verma2025q-cidnet} integrates both technical and aesthetic quality assessment algorithms with the video enhancement task. Building upon the HVI-CIDNet~\cite{yan2025hvi} framework, they introduce a perceptual quality-aware color and intensity decoupling network that leverages the Lighten Cross-Attention (LCA) mechanism. In addition, they incorporate a quality loss function based on CLIP-IQA metrics to enhance the perceptual quality of the output video, ensuring alignment with human visual preferences.

\begin{figure}[ht]
    \centering
    \includegraphics[width=0.9\columnwidth]{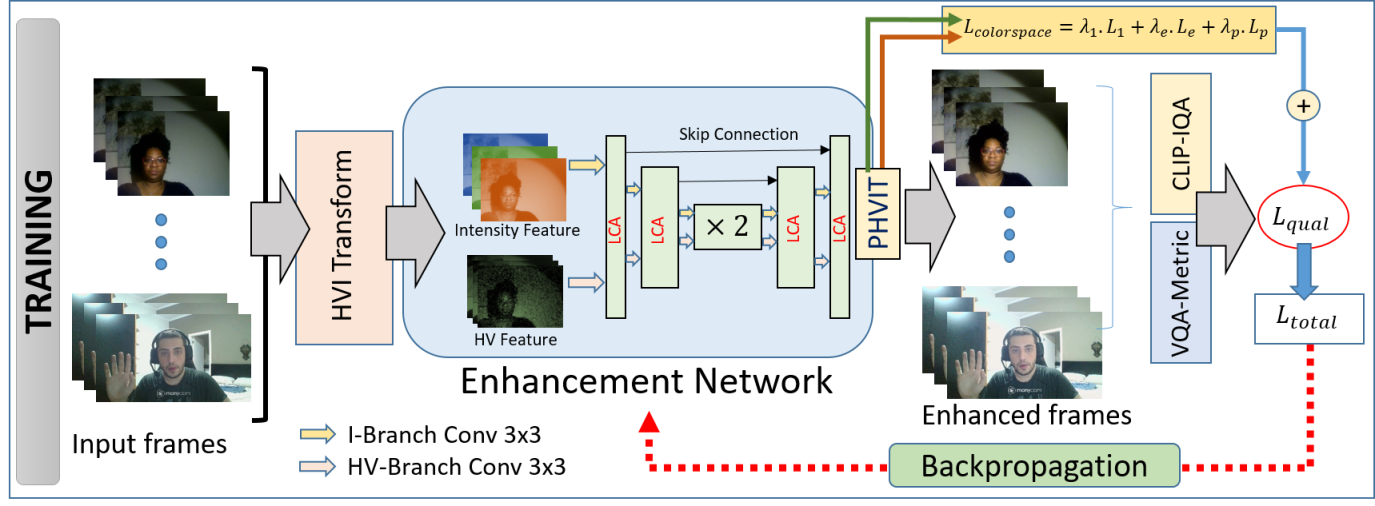}
    \caption{Overview of the quality-aware CIDNet proposed by Wizard. During training, they use extracted frames as input. These inputs are first passed through HVI transform network, the obtained HVI features are then processed in CIDNet, and lastly Perceptual-inverse HVI Transform (PHVIT) is applied to get sRGB-enhanced image. The outputs from the VQE model are evaluated using CLIP-IQA for perceptual quality assessment, and the resulting scores are utilized as a quality loss. The $\mathcal{L}_{colorspace}$ is constructed as combination of L1 Loss ($L_1$), Edge Loss ($L_e$), and Perceptual Loss ($L_p$). }
    \label{fig:model_arch_train}
\end{figure}

To accelerate convergence and enhance performance, they initialize the model with pretrained HVI-CIDNet~\cite{yan2025hvi} weights, leveraging prior knowledge for effective spatial and chromatic feature handling in video frames.
A major limitation of traditional image enhancement models is that they often prioritize technical fidelity over perceptual quality. However, in video conferencing applications, visual appeal is just as important as technical accuracy. To address this, they introduce a perceptual quality loss function that incorporates metrics from CLIP-IQA~\cite{clipiqa}, in combination with the Video Quality Assessment metrics (Equation~\eqref{eq:vqa}). The quality loss component encourages the model to prioritize aesthetic factors, ensuring that the output video frames align closely with human visual preferences.

As given in the block diagram \figurename{~\ref{fig:model_arch_train}}, during training, let \(x_i\) be the input frames and \(\hat{x}_i\) be the enhanced frames generated by the enhancement network. Enhanced frames are fed into the CLIP-IQA model, which computes the quality scores \(Q_{\hat{x}_i}\). Scores typically range between $[0 - 1]$, where a higher score indicates better perceptual quality, the objective is to maximize the score. To incorporate this into the loss function while ensuring optimization, the mean quality score $\bar{Q}(\hat{x}_i)$ across a batch is normalized as follows:

\begin{equation}
    \mathcal{L}_{quality}^{clip} = 1 - \bar{Q_c}(\hat{x}_i),
\end{equation}
and $\bar{Q_c}(\hat{x}_i)$ is given as,
\begin{equation}
    \bar{Q_c}(\hat{x}_i) = \frac{1}{N} \sum_{i=1}^N Q_c(\hat{x}_i),
\end{equation}
where, $N$ represents the batch size. Here, a value of $1$ corresponds to the highest quality, in the same manner, another term from VQA model (Equation~\eqref{eq:vqa}) is constructed as:
\begin{equation}
    \mathcal{L}_{quality}^{VQA} = 1 - \bar{Q_v}(\hat{x}_i)    
\end{equation}
Finally, quality loss is given as:
\begin{equation}
    \mathcal{L}_{quality} = \mathcal{L}_{quality}^{clip} + \mathcal{L}_{quality}^{VQA}
\end{equation}
The term $\mathcal{L}_{quality}$ is then integrated into the overall loss function to guide the training process. The total loss function used during training is a weighted sum of the standard colorspace loss and the perceptual quality loss  :

\begin{equation}
    \mathcal{L}_{total} = \mathcal{L}_{colorspace} + \lambda \cdot \mathcal{L}_{quality}
\end{equation}

where $\mathcal{L}_{colorspace}$ is the loss term used in basemodel~\cite{yan2025hvi}. The hyperparameter $\lambda$ controls the contribution of the perceptual quality term, ensuring a balance between technical accuracy and visual quality.

\begin{figure}[ht]
    \centering
    \includegraphics[width=0.9\columnwidth]{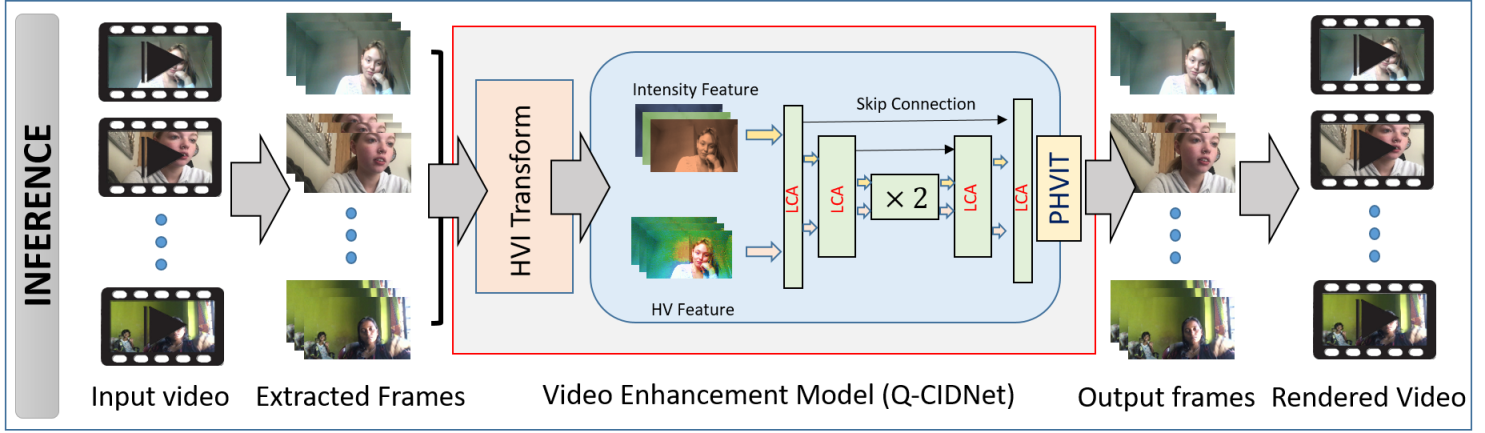}
    \caption{Inference pipeline of proposed quality-aware CIDNet. First the frames are extracted from input video, extracted frames are fed to Q-CIDNet for enhancement.
    }
    \label{fig:model_arch_infer}
\end{figure}

\subsubsection{Datasets}
To develop and evaluate the Video Quality Enhancement (VQE) model, they selectively utilized subsets of the real and synthetic datasets provided in this challenge. Out of the total $10,000$ real videos and $1,500$ synthetic videos, they opted for a focused approach to training by leveraging representative samples from each dataset:
\begin{itemize}
    \item Real Dataset Utilization: From the $10,000$ real videos provided for training and validation, they selected $300$ videos for training. This subset was chosen to capture diverse lighting conditions, variations in ambient reflections, and noise characteristics while maintaining a balance between complexity and model training efficiency.
    \item Synthetic Dataset Utilization: From the $1,500$ synthetic videos provided for training, they selected $300$ videos for training. These synthetic samples were curated to include a variety of lighting configurations generated by adding diffuse light sources to simulate a studio setup. This data was instrumental in fine-tuning the model's ability to handle lighting corrections and improve visual appeal.
\end{itemize}

\subsubsection{Experiments \& Results}
After initializing the model with pretrained weights from HVI-CIDNet, they fine-tuned it using the Adam optimizer with hyperparameters $\beta_1 = 0.9$ and $\beta_2 = 0.999$ for $50$ epochs. The parameter $\lambda$ is set to $0.75$ for quality loss inclusion. The learning rate was initially set to $1 \times 10^{-4}$ and gradually decreased to $1 \times 10^{-7}$ using a cosine annealing schedule during the training process.
 On an input of $3 \times 720 \times 1280$, the model requires $114.249$ GMACs, equivalent to $228.498$ GFLOPs, with $1.973$M parameters.
 The measured inference latency is $170$ ms per frame. Details are listed in Table~\ref{table:results} and inference detailed in \figurename{~\ref{fig:model_arch_infer}}.


    \section*{Acknowledgments}
\label{sec:acknowledgments}

The authors thank the Human Understanding Toolkit team, led by Tadas Baltrusaitis, for their support in using their synthetic data generation framework and adapting it to our needs. We especially thank Lohit Petikam for his critical help in designing the studio lighting setup in Blender and for collaborating on writing the rendering scripts.
This work was partially supported by the Humboldt Foundation. We thank the NTIRE 2025 sponsors: ByteDance, Meituan, Kuaishou, and University of Wurzburg (Computer Vision Lab).

    \appendix
    \section{Teams \& Affiliations}
\label{sec:appendix_teams}

\subsection*{NTIRE 2025 Team}
\noindent\textit{\textbf{Title:}} NTIRE 2025 Challenge on Video Quality Enhancement for Video Conferencing\\
\noindent\textit{\textbf{Members:}}\\
Varun Jain$^1$ (\href{mailto:jain.varun@microsoft.com}{jain.varun@microsoft.com}),\\
Zongwei Wu$^2$ (\href{mailto:zongwei.wu@uni-wuerzburg.de}{zongwei.wu@uni-wuerzburg.de}),\\
Quan Zou$^1$ (\href{mailto:quan.zou@microsoft.com}{quan.zou@microsoft.com}),\\
Louis Florentin$^1$ (\href{mailto:lflorentin@microsoft.com}{lflorentin@microsoft.com}),\\
Henrik Turbell$^1$ (\href{mailto:heturbel@microsoft.com}{heturbel@microsoft.com}),\\
Sandeep Siddhartha$^1$ (\href{mailto:ssiddhartha@microsoft.com}{ssiddhartha@microsoft.com}),\\
Radu Timofte$^2$ (\href{mailto:radu.timofte@uni-wuerzburg.de}{radu.timofte@uni-wuerzburg.de})\\
\noindent\textit{\textbf{Affiliations:}}\\
$^1$ Microsoft, Redmond WA, USA\\
$^2$ Computer Vision Lab, University of W\"urzburg, Germany\\

\subsection*{TMobileRestore}
\noindent\textit{\textbf{Members:}}\\
Qifan Gao$^1$ (\href{mailto:qf\_gao@outlook.com}{qf\_gao@outlook.com}),\\
Linyan Jiang$^1$ (\href{mailto:jly724215288@gmail.com}{jly724215288@gmail.com}),\\
Qing Luo$^1$ (\href{mailto:luoqing.94@qq.com}{luoqing.94@qq.com}),\\
Jie Song$^2$ (\href{mailto:553252129sj@gmail.com}{553252129sj@gmail.com}),\\
Yaqing Li$^1$ (\href{mailto:lyqstudy@qq.com}{lyqstudy@qq.com})\\
\noindent\textit{\textbf{Affiliations:}}\\
$^1$ ShannonLab, Tencent, China\\
$^2$ Xidian University, China\\

\subsection*{Summer}
\noindent\textit{\textbf{Members:}}\\
Summer Luo$^1$ (\href{mailto:185471613@qq.com}{185471613@qq.com}),\\
Mae Chen$^2$ (\href{mailto:chenxm\_m@nankai.edu.cn}{chenxm\_m@nankai.edu.cn})\\
\noindent\textit{\textbf{Affiliations:}}\\
$^1$ Independent researcher, China\\
$^2$ Nankai Uninversity, China\\

\subsection*{DeepView}
\noindent\textit{\textbf{Members:}}\\
Stefan Liu$^1$ (\href{mailto:stefan1026@163.com}{stefan1026@163.com}),\\
Danie Song$^2$ (\href{mailto:hypox128@gmail.com}{hypox128@gmail.com})\\
\noindent\textit{\textbf{Affiliations:}}\\
$^1$ Shanghai Jiao Tong University, China\\
$^2$ Shenzhen University, China\\

\subsection*{Maqic}
\noindent\textit{\textbf{Members:}}\\
Huimin Zeng$^1$ (\href{mailto:zeng.huim@northeastern.edu}{zeng.huim@northeastern.edu}),\\
Qi Chen$^2$ (\href{mailto:qchen76@jh.edu}{qchen76@jh.edu})\\
\noindent\textit{\textbf{Affiliations:}}\\
$^1$ Northeastern University, USA\\
$^2$ Johns Hopkins University, USA\\

\subsection*{Wizard}
\noindent\textit{\textbf{Title:}} Q-CIDNet: Perceptual Quality Aware Color and Intensity Decoupling Network for Video Quality Enhancement\\
\noindent\textit{\textbf{Members:}}\\
Ajeet Kumar Verma$^1$ (\href{mailto:ajeet.verma@iitjammu.ac.in}{ajeet.verma@iitjammu.ac.in}),\\
Shweta Tripathi$^1$ (\href{mailto:2023pcs0041@iitjammu.ac.in}{2023pcs0041@iitjammu.ac.in}),\\
Vinit Jakhetiya$^1$ (\href{mailto:vinit.jakhetiya@iitjammu.ac.in}{vinit.jakhetiya@iitjammu.ac.in}),\\
Badri N Subhdhi$^2$ (\href{mailto:subudhi.badri@iitjammu.ac.in}{subudhi.badri@iitjammu.ac.in}),\\
Sunil Jaiswal$^3$ (\href{mailto:sunil.jaiswal@k-lens.de}{sunil.jaiswal@k-lens.de})\\
\noindent\textit{\textbf{Affiliations:}}\\
$^{1}$Department of Computer Science and Engineering, IIT Jammu, India\\
$^{2}$Department of Electrical Engineering, IIT Jammu, India\\
$^{3}$K$|$Lens, GmbH, Germany\\

    {
        \small
        \bibliographystyle{ieeenat_fullname}
        \bibliography{main}
    }
\end{document}